\documentclass{article}

\PassOptionsToPackage{numbers, compress}{natbib}

\usepackage[preprint]{neurips_2026}

\usepackage{wrapfig}
\newcommand{\XComment}[1]{}

\usepackage[utf8]{inputenc} %
\usepackage[T1]{fontenc}    %
\usepackage{hyperref}       %
\usepackage{url}            %
\usepackage{booktabs}       %
\usepackage{amsfonts}       %
\usepackage{amsmath}        %
\usepackage{algorithm}      %
\usepackage{algpseudocode}  %
\usepackage{nicefrac}       %
\usepackage{microtype}      %
\usepackage{xcolor}         %

\usepackage{enumitem}
\usepackage{graphicx}
\usepackage{float}

\title{\NAME{}: Efficient MoE LLM Serving with Dynamic Quality-Aware Weight Quantization}

\author{%
  Yuchen Yang$^{1}$ \quad
  Yifan Zhao$^{1}$ \quad
  Anisha Dasgupta$^{1}$ \quad
  Sasa Misailovic$^{1}$ \\
  $^{1}$University of Illinois Urbana-Champaign \\
  \texttt{\{yucheny8, yifanz16, adasg, misailo\}@illinois.edu}
}

\def\NAME{PagedWeight}
\def\COMMENTSON{true}  

\ifdefined\COMMENTSON
\definecolor{mauve}{HTML}{9558B2}

\newcommand{\sasa}[1]{\textcolor{blue}{\bf Sasa: #1}}
\newcommand{\yuchen}[1]{\textcolor{orange}{Yuchen: #1}}
\newcommand{\anisha}[1]{\textcolor{violet}{Anisha: #1}}
\newcommand{\yifan}[1]{\textcolor{mauve}{Yifan: #1}}
\else

\newcommand{\sasa}[1]{}
\newcommand{\yuchen}[1]{}
\newcommand{\anisha}[1]{}
\newcommand{\yifan}[1]{}
\fi

\begin{document}

\maketitle

\begin{abstract}
  
Mixture-of-Experts (MoE) is a popular class of large language models (LLMs), offering high efficiency and accuracy. However, in KV-cache-intensive serving scenarios, MoEs often exhibit a tension between the GPU memory requirements of the model weights and the growing KV cache. 
We propose \NAME{}, a novel management method for MoE LLM serving that
dynamically quantizes MoE model's weights at runtime and balances expert-weight precision with the KV cache sizes. 
\NAME{} exposes and effectively navigates the complex tradeoff between the model's task accuracy, memory consumption, and throughput/latency. 
Across several memory-sensitive MoE serving scenarios, \NAME{} improves the quality-memory tradeoff over several existing quantization baselines. \NAME{}
achieves FP16-equivalent accuracy with up to 72.0\% GPU memory savings and 1.94$\times$ throughput improvement, and improves quality over quantization methods by up to 39.3\% at a similar memory budget with at most \mbox{4.1\% throughput loss.}

\end{abstract}

\section{Introduction}

Mixture-of-Experts (MoE) models are gaining traction for efficiently solving complex language
and reasoning tasks.
An MoE layer consists of a bank of expert networks
and a router that selects a subset of experts that are likely to perform well.
Since only a subset of experts is active,
MoE significantly reduces the amount of computation
without sacrificing accuracy~\cite{gshard,switch,mixtral},
making it particularly appealing for long-context tasks,
such as repository-level coding and long document analysis.

However, MoE scales up the total parameter count and leaves a large memory footprint
on the GPU.
GPU memory management is a key challenge in modern LLM serving systems,
which need to store both the model weights and the KV cache in the GPU memory during inference.
The size of KV caches grows with the context length and is a main source of memory pressure, for which system developers have developed various techniques to mitigate this.
For example, vLLM PagedAttention manages the KV cache in paged blocks
to reduce fragmentation and increase batching efficiency~\cite{vllm},
and KV cache compression reduces memory usage with a minimal impact on accuracy~\cite{kvcomp}.

Yet, all of these methods are limited by the other significant memory bottleneck,
which is MoE weights in the GPU memory.
With MoE, the loaded model size can occupy more than 60\% of the GPU memory~\cite{vllm}.
A convenient and automatic approach to reduce the memory footprint of MoE models
can create more space for the KV cache and allow for longer contexts.
Moreover, it opens up an interesting and unexplored tradeoff space
between the task accuracy, computation time,
and the allocation and balancing of GPU memory for model weights and the KV cache. 

Quantization is the most common way to reduce model size,
as demonstrated by many post-training techniques~\cite{gptq,smoothquant,awq,spqr}.
However, most quantization techniques can only be applied statically,
i.e., they only modify weights before the model runs. 
Existing runtime precision-adaptation methods show that
precision does not need to be fixed for an entire request~\cite{dpllm}. 
However, they do not use precision changes to reallocate GPU memory at runtime. In MoE serving, a key challenge is to free memory from less critical expert weights as the KV cache grows, without hurting quality or interrupting inference.

\begin{figure}
    \vspace{-1.4em}
    \centering

    \includegraphics[width=0.9\textwidth]{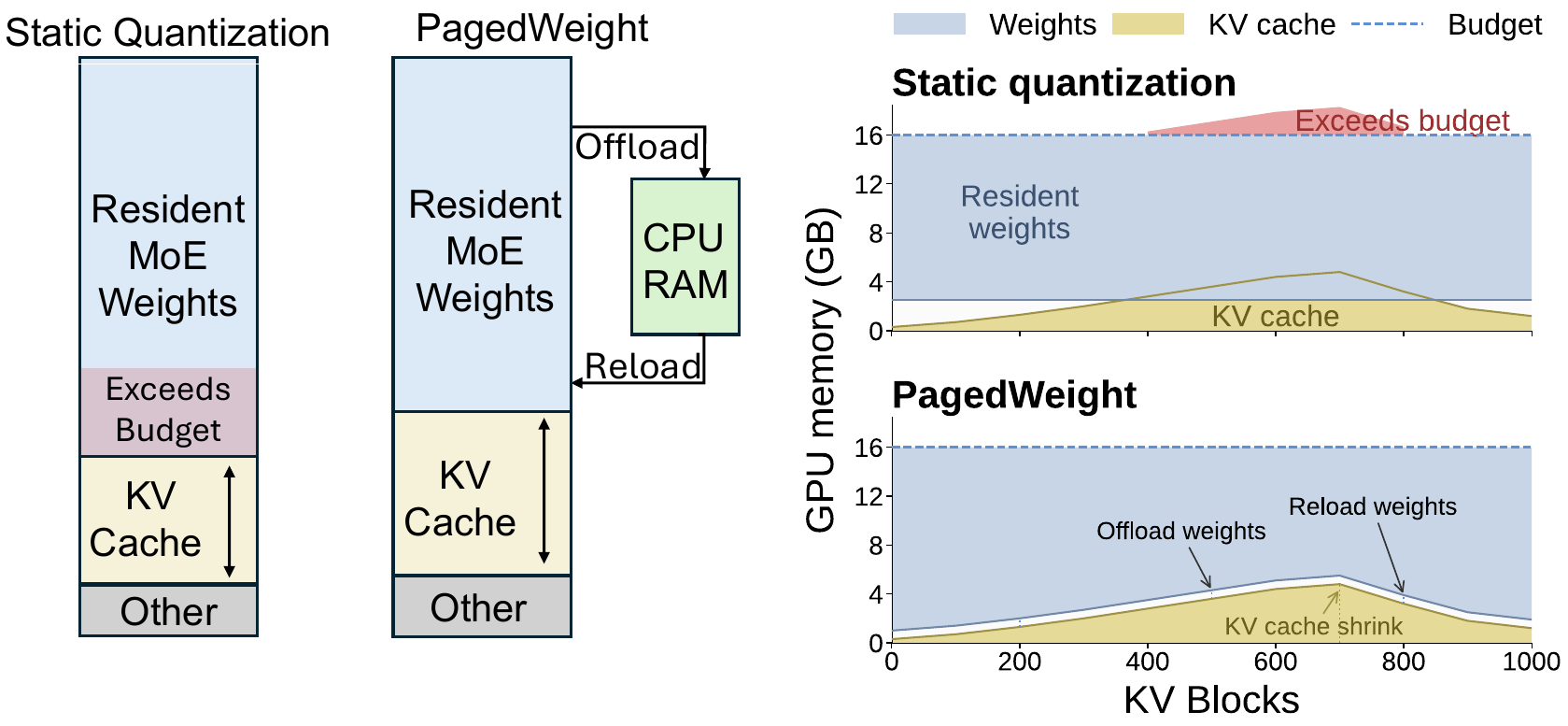}

    \caption{Static quantization fixes MoE weights; PagedWeight offloads weight pages to free KV cache headroom.}
    \label{fig:motivation}
    \vspace{-.2in}
\end{figure}

\noindent\textbf{Our Work:}
We propose \NAME{}, a novel management method for MoE LLM serving that dynamically quantizes MoE model's weights at runtime and balances the choice of quantization with the KV cache and context sizes. \NAME{} builds on our insight that experts' weights can be treated in a similar way as paged blocks in KV caches, thus avoiding inflexibility of static quantization (see Figure~\ref{fig:motivation}). We designed \NAME{} to satisfy the following three objectives: 
\vspace{-.08in}
\begin{itemize}[leftmargin=.1in]\itemsep 1pt\parskip 1pt
  \item \textbf{Dynamic KV/weight tradeoff.} \NAME{} should release
  GPU memory that model weights consume when the KV cache grows and needs more GPU memory.
  \item \textbf{Quality-aware expert selection.} \NAME{} should choose MoE linear-blocks for which it predicts that the bitwidth changes will have the smallest impact on the request's output quality.
  \item \textbf{Asynchronous execution.} Page movement should overlap
  with the model serving and should not add extra cost.
\end{itemize}
\vspace{-.08in}

To effectively represent multiple quantized experts, 
\NAME{} leverages Any-Precision LLM (APL), which can represent multiple bitwidths in an overlaid bit-plane format~\cite{apl}.
\NAME{} treats the Any-Precision (AP) bit-plane and lookup table (LUT)
buffers of routed expert linear-blocks as GPU-resident pages whose bitwidth
can be lowered or restored at runtime. Unlike PagedAttention, which pages KV
cache blocks, \NAME{} manages quantized weight state. Unlike static MoE
quantization, it adapts the expert precision to current KV cache size. 

\NAME{}'s structured planner uses both offline and runtime analyses to determine the quantization level that is likely not to impact task accuracy. Those include offline sensitivity (calibrated Hessian-weighted quality-damage scores),
routing-aware grouping (protection for frequently routed experts), and prompt residuals (prompt-specific corrections to quantization-impact estimates). To optimize execution time and hide memory transfer latencies, \NAME{} offloads/reloads quantized weight pages asynchronously and
commits only at safe boundaries. 
We also
implement a fused mixed-precision MoE kernel that reads Any-Precision bit-planes and
LUTs directly.

We evaluate \NAME{} on three common open MoE models, ranging in size between 14.3B and 46.7B and against three state-of-the-art quantization methods - APL~\cite{apl}, DP-LLM~\cite{dpllm}, and MxMoE~\cite{mxmoe}. \NAME{}
consistently achieves a better task quality-memory tradeoff than all evaluated
quantization baselines. \NAME{}
achieves FP16-equivalent accuracy with up to 72.0\% GPU memory savings and 1.94$\times$ throughput improvement, and improves
quality over quantization methods by up to 39.3\% at a similar memory budget with at most 4.1\%
  throughput loss.

\noindent\textbf{Contributions.} This paper makes the following contributions:
\begin{itemize}[leftmargin=.1in]\itemsep 2pt
\vspace{-.1in}

  \item We propose \NAME{}, a paged-weight system for online Any-Precision MoE
  weights that manages the committed bitwidths, desired bitwidths, page states, and 
  memory consumptions.

  \item We design a quality-aware runtime planner that combines offline sensitivity,
  online routing statistics, and prompt residuals to choose a low-damage 
  movement strategy for the weight page under an increased memory demand from KV cache.

  \item We use asynchronous page movement to hide offload/reload latency,
  with \mbox{minimal throughput loss.}
  
  \item We evaluate \NAME{} on three popular MoE models and demonstrate that it outperforms all evaluated quantization baselines.
  \vspace{-.1in}
\end{itemize}

\section{Background}

\subsection{Mixture-of-Experts Architecture and Routing Imbalance}
\label{sec:bg-moe-routing}

In an MoE layer, the router assigns routing weights to \(E\) experts and
selects the top \(K\) experts for each token~\cite{mixtral}.
During inference, experts in a layer are selected with highly unequal frequency~\cite{switch}.
This \emph{routing imbalance} indicates that experts are not all equally important,
and locating the less important experts allows us to quantize them more aggressively
with minimal accuracy loss.
\begin{figure}
  \vspace{-0.1in}
  \centering
  \includegraphics[width=.66\linewidth]{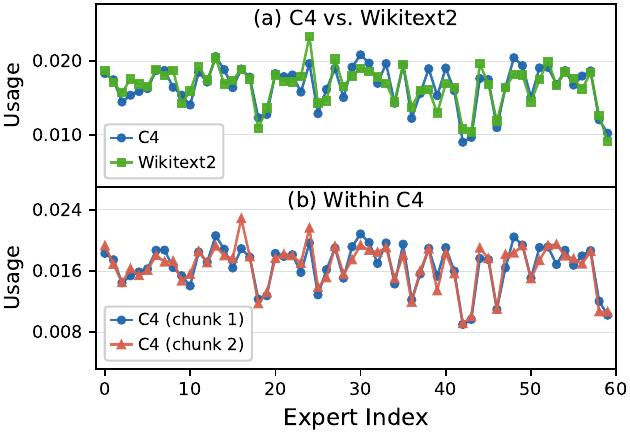}
  \vspace{-0.1in}
  \caption{Expert activation patterns on Qwen1.5-MoE-A2.7B.
  The y-axis shows the normalized routing weight assigned to each expert. }
  \label{fig:expert_activation}
\end{figure}
The \emph{routing mass} of an expert captures its importance,
which is intuitively its routing weight normalized by the total weight of all experts in a layer.
It is formally defined, for an expert $e$ in layer $\ell$, as:
\begin{equation}
  m_{\ell,e} = \frac{
  \sum_t \sum_k r_{\ell,t,k}\mathbf{1}\{z_{\ell,t,k}=e\}}{
  \sum_t \sum_k r_{\ell,t,k}}
\end{equation}
$t$ indexes observed tokens, $k$ indexes the top-$k$ router choices,
$z_{\ell,t,k}$ is the selected expert, and $r_{\ell,t,k}$ is its routing weight.
As shown in Figure~\ref{fig:expert_activation}, routing mass varies substantially across
experts and inputs.

\subsection{Mixed-Precision Quantization and Runtime Quantization}
\label{sec:bg-apweights}

To achieve higher accuracy, mixed-precision quantization (MPQ)
assigns a different bitwidth to each component of the network (such as a weight matrix or an expert),
and often uses the component's \emph{sensitivity} to estimate the accuracy impact.
Hessian-based sensitivity is a common approach used in previous work~\citep{gptq,dpllm},
which defines the sensitivity of a weight matrix $W_{i,j}$ at bitwidth $b$ as:
\begin{equation}
  s_i^b = \sum_j h_{i,j} \left( W_{i,j} - W_{i,j}^b \right)^2
\end{equation}
where \(i\) indexes a quantized component, \(j\) indexes elements in weight matrix \(i\), \(h_{i,j}\) is a Hessian-derived importance score of $W_{i,j}$,
and \(W_{i,j}^b\) is the \(b\)-bit quantized version of \(W_{i,j}\).
Lower \(s_i^b\) indicates that the weight can tolerate more aggressive quantization.
In the context of MoE models, each expert typically consists of weight \textit{linear-blocks}~\cite{mxmoe}.
Within the same expert, linear-blocks can have different sensitivity to quantization,
and quantizing them to different bitwidths can yield better accuracy.

Most existing quantization methods produce a fixed quantized model offline,
which does not change at serving time~\citep{awq, gptq, smoothquant, spqr},
thus missing out on the opportunity to adapt to dynamic inference conditions.
There are a few works that explore \emph{runtime quantization} of LLMs,
such as APL~\citep{apl},
which allows the quantization level of the model to change at inference time.

APL stores weights in a shared \emph{bit-plane}
format, where different subsets of bit-planes correspond to different
effective bitwidths. Each quantized tensor is also associated with a
lookup table (LUT), which stores the bitwidth-specific centroid values used by the AP representation.

\section{PagedWeight System}
\label{sec:method}

\begin{figure}[t]
    \centering
    \includegraphics[width=\linewidth]{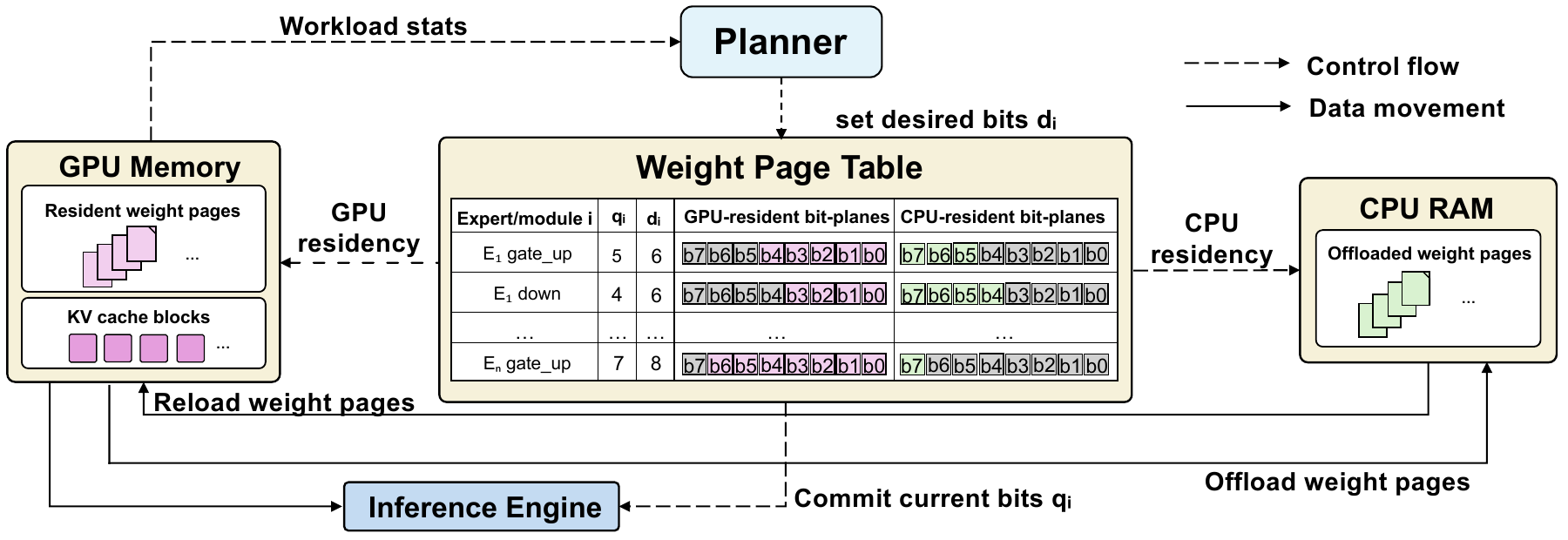}

    \caption{\textbf{\NAME{} system overview.}
    Dashed arrows show control flow, and solid arrows show weight data movement.}
    \label{fig:pagedweights-overview}
    \vspace{-.2in}
\end{figure}

We illustrate the workflow of \NAME{} in Figure~\ref{fig:pagedweights-overview}.
The \emph{weight page table} records for each linear-block (of each expert in each MoE layer)
the desired bitwidth $d_i$ and the committed bitwidth $q_i$.
$q_i$ is the precision currently used in inference.
The \emph{planner} is \NAME{}'s runtime controller, in charge of quantization decisions
and data movements.
The planner decides desired bitwidths $d_i$ and updates them in the weight page table,
based on the current KV-cache pressure, routing statistics, and prompt-specific features.
\NAME{} executes planned quantization decisions (making actual $q_i$ match planned $d_i$)
by moving weight pages between CPU and GPU, when it is safe to update them.

\subsection{Paged Weight Representation}
\label{sec:method-paged-weights}

\NAME{} quantizes a network at the granularity of per expert linear-block (a weight in an expert).
Each linear-block is identified by a triple index: $i := (\ell,e,u)$
for layer $\ell$, expert $e$, and one of two linear-block types
$u\in\{\texttt{gate\_up},\texttt{down}\}$.
\NAME{} organizes linear-blocks and their quantization decision in a weight page table,
where each linear-block $i$ occupies one row.
The page table row for $i$ stores:
a set of supported bitwidths $\mathcal{B}_i$ for the planner to choose from,
the committed bitwidth $q_i$ that is currently used in inference,
and the desired bitwidth $d_i$ selected by the planner.
Since linear-block weights are stored in a bit-plane representation,
each row also carries a lookup table for the state of each bit-plane,
where the state can be GPU-resident, CPU-resident, or in transfer.
One bit-plane and its lookup table state together is one \emph{\textbf{weight page}}.

Each page table row $i$ also contains a GPU-resident memory size table $M_i^b$
for each $b\in\mathcal{B}_i$, which records memory size for weight of each precision.
If a linear-block weight goes from bitwidth $b$ to a lower bitwidth $b'$,
the amount of GPU memory released, $\Delta_i^{b\to b'}$, is
\begin{equation}
  \Delta_i^{b\to b'} = M_i^b - M_i^{b'} \qquad b'<b
\end{equation}
$M_i^b$ includes both the AP bit-plane and LUT for $b$-bit of linear-block $i$.

\NAME{} moves data between CPU and GPU to carry out planned quantization decisions.
During an \emph{offload}, \NAME{} first reduces the committed bitwidth $q_i$ in the table,
and then moves unneeded pages from GPU memory to CPU memory.
During a \emph{reload}, \NAME{} first moves pages from CPU memory back to GPU memory.
It only raises $q_i$ when the page state becomes GPU-resident.

\subsection{Quality-Aware Runtime Planner}
\label{sec:method-planner}

The runtime planner of \NAME{} makes quantization decisions based on the
estimated impact of bitwidth reductions on model accuracy. A candidate action
considered by the planner is $(i,b\to b')$, where $i=(\ell,e,u)$ and $b'<b$.
For each action, the planner estimates accuracy impact by combining three
factors: offline global sensitivity, online routing statistics, and runtime
prompt residual.

As discussed in Section~\ref{sec:bg-apweights}, offline sensitivity can guide bitwidth assignment with budget. \NAME{} uses this assignment once before
serving to initialize committed bitwidths $q_i$. During serving, the planner
refreshes plans using runtime observations. Online \emph{routing statistics} identify
frequently selected experts and increase the estimated impact of quantizing
their linear-blocks, protecting hot experts. Runtime \emph{prompt residual}
provides a prompt-conditioned correction to the offline sensitivity estimate,
capturing that the same quantization action can have different impact on
different prompts.

\begin{wrapfigure}{R}{0.55\textwidth}
\vspace{-.38in}
\begin{minipage}{0.55\textwidth}
\begin{algorithm}[H]
\small
\caption{\NAME{} Runtime Planner}
\label{alg:pagedweight-planner}
\textbf{Inputs:} $\mathcal{T}_t$: page table state, $\mathcal{O}_t$: runtime observation, $\Omega_{\mathrm{off}}$: offline calibration results, $Q_t$: current plan queue.\\
\textbf{Return:} $Q_{t+1}$: updated plan queue, $\Pi_t$: page movement plan $\{(i,q_i\!\rightarrow\!d_i,\Delta_i)\}$.
\begin{algorithmic}[1]
\Function{PlanStep}{$\mathcal{T}_t,\mathcal{O}_t,\Omega_{\mathrm{off}},Q_t$}\label{algline:planstep}
\State $Q_{t+1} \gets Q_t$\label{algline:init-queue}
\State $\Pi_t \gets \emptyset$\label{algline:init-plan}
\State $D_t \gets \textsc{TargetBytesFromKVPressure}(\mathcal{O}_t)$\label{algline:target-bytes}
\ForAll{$s \in \textsc{BitwidthFloorStages}(\Omega_{\mathrm{off}})$}\label{algline:for-cap-stage}
    \State $\mathcal{R}_{s} \gets \textsc{ScoreActions}(\mathcal{T}_t,\mathcal{O}_t,\Omega_{\mathrm{off}},s)$\label{algline:score-actions}
    \State $\Pi_{s} \gets \textsc{GreedySelect}(\mathcal{R}_{s},D_t)$\label{algline:greedy-select}
    \State $Q_{t+1} \gets \textsc{RefreshPlanQueue}(Q_{t+1},\Pi_{s})$\label{algline:refresh-queue}
\EndFor\label{algline:end-cap-stage}
\If{$\textsc{PressureTriggered}(\mathcal{O}_t)$}\label{algline:pressure-triggered}
    \State $\Pi_t \gets \textsc{SelectBestPlan}(Q_{t+1},D_t)$\label{algline:return-plan}
\EndIf\label{algline:end-if}
\State \Return $(Q_{t+1},\Pi_t)$\label{algline:return-step}
\EndFunction\label{algline:end-function}
\end{algorithmic}
\end{algorithm}
\end{minipage}
\vspace{-.2in}
\end{wrapfigure}

Algorithm~\ref{alg:pagedweight-planner} shows one step of \NAME{}'s runtime
planner. The input $\mathcal{T}_t$ is the page-table state at time step $t$,
which includes the committed bitwidth $q_i$, desired bitwidth $d_i$, page
states, and weight size table $M_i^b$ for linear-block $i$ at bitwidth $b$.
$\mathcal{O}_t$ is the runtime observation at $t$, including KV-cache
pressure, routing statistics, and prompt features. $\Omega_{\mathrm{off}}$
stores offline calibration results, including sensitivity scores,
prompt-residual heads, and bitwidth floor. The planner also takes the current
plan queue $Q_t$ as input and returns an updated $Q_{t+1}$. If memory
pressure is triggered, it also returns a page-movement plan $\Pi_t$.

The planner step first copies $Q_{t}$ to $Q_{t+1}$ and initializes the plan $\Pi_t$
(lines~\ref{algline:init-queue}--\ref{algline:init-plan}). It then converts
KV-cache pressure into the byte target $D_t$
(line~\ref{algline:target-bytes}). For each bitwidth-floor stage $s$ from
$\Omega_{\mathrm{off}}$, \textsc{ScoreActions} builds the legal transition set
$\mathcal{R}_{s}$, \textsc{GreedySelect} selects a candidate plan
$\Pi_{s}$ to meet $D_t$, and \textsc{RefreshPlanQueue} updates $Q_{t+1}$
(lines~\ref{algline:for-cap-stage}--\ref{algline:end-cap-stage}). When pressure
exceeds the threshold, \textsc{SelectBestPlan} selects the best plan from
$Q_{t+1}$ as $\Pi_t$
(lines~\ref{algline:pressure-triggered}--\ref{algline:return-step}). The runtime
invokes this planner step repeatedly as the request runs, so later steps can
refresh the queue with newer observations.

\noindent\textbf{Offline global sensitivity.}
As discussed in Section~\ref{sec:bg-apweights}, offline calibration gives a
sensitivity score $s_i^b$ for each linear-block $i$ and supported bitwidth
$b\in\mathcal{B}_i$. \NAME{} reuses these scores and defines the global damage of
reducing the bitwidth of linear-block $i$ from $b$ to $b'$ as
\begin{equation}
  g_i^{b\to b'} =
  \max\{s_i^{b'}-s_i^b,0\}  \qquad  b' < b
\end{equation}

\noindent\textbf{Routing statistics.}
As discussed in Section~\ref{sec:bg-moe-routing}, MoE routing is highly
imbalanced, and the routing mass $m_{\ell,e}$ measures how much traffic an expert
receives within a layer. During serving, \NAME{} collects the frequency of expert
selection and routing weights, and updates routing mass online.

\NAME{} sorts experts in each layer by their routing mass and assigns them to
different buckets. $\beta_{\ell,e}$ is the bucket of expert $e$ in layer $\ell$. 
Each bucket has a
damage multiplier $\mu_{\beta_{\ell,e}}$ and a bitwidth floor. Hotter
buckets contain experts with larger routing mass. They use larger damage
multipliers and a higher bitwidth floor to protect them. Linear-blocks of an
expert share the same routing bucket. 

\noindent\textbf{Prompt residual.}
\NAME{} also uses \emph{prompt residual}, which is a prompt-sensitive
correction to the global damage metric. Prompt residual is useful because
quantization sensitivity depends on the input. We keep the global damage as a prompt-independent prior, and use the prompt residual to
adjust this estimate for the current input. The residual is predicted from
prompt-specific features using linear regression heads. Each head is specific to a linear-block type and
bitwidth transition $(b\to b')$, and is shared across linear-blocks of that type.

The global damage $g_i^{b\to b'}$ estimates the average damage of
reducing the bitwidth of linear-block $i$ from $b$ to $b'$, but an action can have
different damage for different inputs. \NAME{} therefore adds a prompt
residual on top of the global damage.

In offline calibration, we measure the per-sequence damage for each candidate
action $(i,b\to b')$ and compare it with the global estimate. The training target
is their log residual $\rho_i^{b\to b'}$, where a positive value means the prompt is more sensitive
than the global estimate and a negative value means it is less sensitive. 
We train a linear regression head for each linear-block type $u$ and bitwidth transition
$b\to b'$:
\begin{equation}
  \widehat{\rho}_i^{b\to b'} =
  (w_u^{b\to b'})^\top \phi_i + a_u^{b\to b'} .
\end{equation}
Here, $\phi_i$ is a three-dimensional routed-input feature vector: routing-weighted
mean input norm, routing-weighted RMS input norm, and maximum input norm.

At runtime, \NAME{} computes $\phi_i$ for the current sequence and uses the
trained linear head to predict the log residual $\widehat{\rho}_i^{b\to b'}$.
The prediction is clipped and scaled by a confidence score $c_u^{b\to b'}$ and
a strength hyperparameter $\alpha$:
\begin{equation}
  \eta_i^{b\to b'} =
  \exp\!\left(
  \alpha c_u^{b\to b'}
  \operatorname{clip}\!\left(
  \widehat{\rho}_i^{b\to b'},
  \rho_{\min},
  \rho_{\max}
  \right)\right).
\end{equation}
The planner uses $g_i^{b\to b'} \eta_i^{b\to b'}$ as the prompt-adjusted damage.

\noindent\textbf{Planning objective.}
The hyperparameters in the planner are decided in calibration; more details are in
Section~\ref{sec:methodology}.
The planner quantifies KV-cache pressure as a target number of bytes $D$: 
\begin{equation}
  D = \max\{0,T_{\mathrm{blk}}+1-F_{\mathrm{blk}}\}\,B_{\mathrm{KV}}
\end{equation}
where $T_{\mathrm{blk}}$ is the free-block threshold, $F_{\mathrm{blk}}$ is
the current number of free blocks, and $B_{\mathrm{KV}}$ is the byte size of one
KV block. For each legal candidate action, the predicted damage $\widehat{d}_i^{b\to b'}$ is
\begin{equation}
  \widehat{d}_i^{b\to b'} =
  \max\{\epsilon,\mu_{\beta_i}\,g_i^{b\to b'}\eta_i^{b\to b'}\}
\end{equation}
where $\epsilon$ is a small damage floor, and $\mu_{\beta_i}$ is the damage 
multiplier of the routing bucket of
expert $(\ell,e)$. $g_i^{b\to b'} \eta_i^{b\to b'}$ is the prompt-adjusted damage.

For batched serving, the planner first merges request-level signals into one
batch-level table. 
The routing masses, prompt
residuals, confidence scores, and clipping bounds are averaged 
based on the request's KV block usage
weights. 

The planner considers each legal bitwidth reduction as one step. It sorts the
steps by predicted damage per released byte, $\widehat{d}/\Delta$, and greedily
takes the cheapest steps until the target $D$ is met. Depth
caps are relaxed only when the current caps cannot provide enough bytes. If no
legal plan fully satisfies the target, the planner returns the largest safe
reduction under the current constraints.

\subsection{Asynchronous Page Movement and Kernel Optimization}
\label{sec:method-runtime-integration}

\begin{figure}[t]
    \centering
    
      \includegraphics[width=\linewidth]{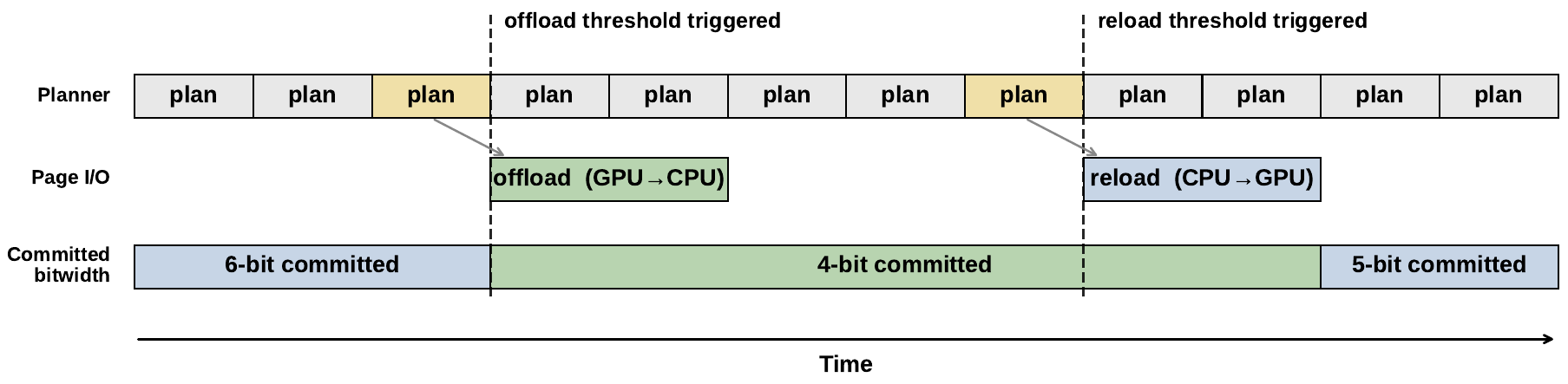}
    
    \caption{\textbf{Asynchronous page-movement pipeline.}
    \NAME{} planner keeps preparing precision-transition plans during runtime. At the offload
    threshold, the latest plan lowers the committed bitwidth and releases weight pages
    from GPU memory. At the reload threshold, the weight pages are restored before
    the higher bitwidth is committed.}
    \vspace{-.2in}
    \label{fig:async-pipeline}
\end{figure}

\textbf{Asynchronous Page Movement.}
Figure~\ref{fig:async-pipeline} illustrates the asynchronous page movement pipeline. During serving, the 
planner monitors free KV blocks, allocator headroom, and
pending KV growth, and keeps generating new plans with the updated information. When pressure exceeds the threshold, the planner chooses the newest plan destination bitwidth for selected linear-blocks.

For offload, the system copies the weight page to CPU RAM and commits the lower $q_i$.
Once the transition is safe, the system retires the GPU pages above the 
committed precision, and the released allocator headroom can be
used by the KV blocks. 
For reload, the runtime restores the offloaded
weight pages first and commits the higher $q_i$ only after the page state is GPU-resident.

\textbf{Fused Mixed-Precision MoE Kernel.}
\NAME{} provides a fused mixed-precision MoE AP kernel to enable per-expert bitwidths at runtime. The kernel reads weight pages directly and enables different 
bitwidths for each linear-block, and does the routing, expert activation, down projection, and output accumulation in one fused CUDA kernel. The kernel fusion reduces the overhead of executing the selected experts
separately. Since the AP state is maintained at linear-block granularity, the
kernel can follow the bitwidth chosen by the planner for each \texttt{gate\_up}
and \texttt{down} module.

\section{Experimental Methodology}
\label{sec:methodology}

\begin{wraptable}[7]{r}{0.5\linewidth}
  \centering
  \footnotesize
  \vspace{-.25in}
  \caption{Architectural specifications of evaluated MoE models.}
  \label{tab:moe-architecture}
  \resizebox{\linewidth}{!}{%
  \begin{tabular}{@{}lccc@{}}
    \toprule
    \textbf{Model Variant} & \textbf{Params (B/GB)} & \textbf{Experts} & \textbf{TopK} \\
    \midrule
    Qwen1.5-MoE-A2.7B & 14.3 / 26.7 & 60+4 & 4 \\
    Mixtral-8$\times$7B-v0.1 & 46.7 / 92.9 & 8 & 2 \\
    Gemma-4-26B-A4B & 25.2 / 53.1 & 128+1 & 8 \\
    \bottomrule
  \end{tabular}
  }
\end{wraptable}

\textbf{Evaluated Models.}
We evaluate \NAME{} on three state-of-the-art base MoE language models
whose architectures cover different expert granularities and routing patterns.
Qwen1.5-MoE-A2.7B~\citep{qwen15moe} combines 60 routed experts with 4 shared
experts and uses top-4 routing.
Mixtral-8$\times$7B-v0.1~\citep{mixtral} uses 8 routed
experts with top-2 routing.
Gemma-4-26B-A4B~\citep{gemma4moe} represents a higher-expert-count design with
128 routed experts, one shared expert, and top-8 routing.
Table~\ref{tab:moe-architecture} summarizes these architectural differences.

\textbf{Evaluation Metrics.}
For perplexity evaluation, we report results on Wikitext2 and C4.
For downstream reasoning, we use GSM8K and MATH-500.
For long-context evaluation, we report the LongBench scores for Passage Retrieval, 
NarrativeQA, and QMSum.
For serving efficiency, we report generation throughput.
For memory-sensitive comparisons, we report peak process-reserved GPU memory.
For quality-memory comparisons in Section~\ref{sec:evaluation-quality-memory}, we use batch size 16
to better reflect a realistic batched serving scenario.

\textbf{Policy Construction.}
The quantization and runtime hyperparameters are built offline on the C4 calibration set. 
We first
form a prompt-independent sensitivity table for each routed expert linear-block and
supported bitwidth transition. Then we train the prompt residual from
routing weighted input norm features, and use a small calibration search to
choose the routing buckets, bucket multipliers, depth caps, and residual weight
used by the online planner.

\textbf{Hardware and Software Setup.}
Qwen1.5-MoE-A2.7B experiments are conducted on an NVIDIA RTX 6000 Ada
workstation.
The other evaluated models are run on an NVIDIA GH200 Grace Hopper system.
We use vLLM v0.20.1 as the serving backend.

\textbf{Baselines.}
We compare \NAME{} against FP16, Any-Precision LLM~\cite{apl} for uniform
quantization, MxMoE~\cite{mxmoe} for static
mixed-precision quantization, and  DP-LLM~\cite{dpllm} for dynamic mixed-precision quantization. Since the original APL implementation does not
support MoE models, we compare against our MoE fused CUDA implementation of the
uniform APL-style baseline.

\vspace{-.1in}

\section{Evaluation}
\label{sec:evaluation}
\vspace{-.1in}

\noindent\textbf{RQ1:}
Can \NAME{} improve the quality-memory tradeoff of MoE LLMs across
different tasks?

\noindent\textbf{RQ2:}
Can \NAME{} preserve long-context quality under large per-request KV-cache demand?

\noindent\textbf{RQ3:}
Can \NAME{} improve model quality while maintaining the throughput?

\noindent\textbf{RQ4:}
Ablation studies: How much do components of \NAME{} contribute to the performance?

\vspace{-.1in}

\subsection{Quality-Memory Tradeoff}
\label{sec:evaluation-quality-memory}

\begin{figure}[th]
\centering
\hspace*{-.22in}\includegraphics[width=1.06\linewidth]{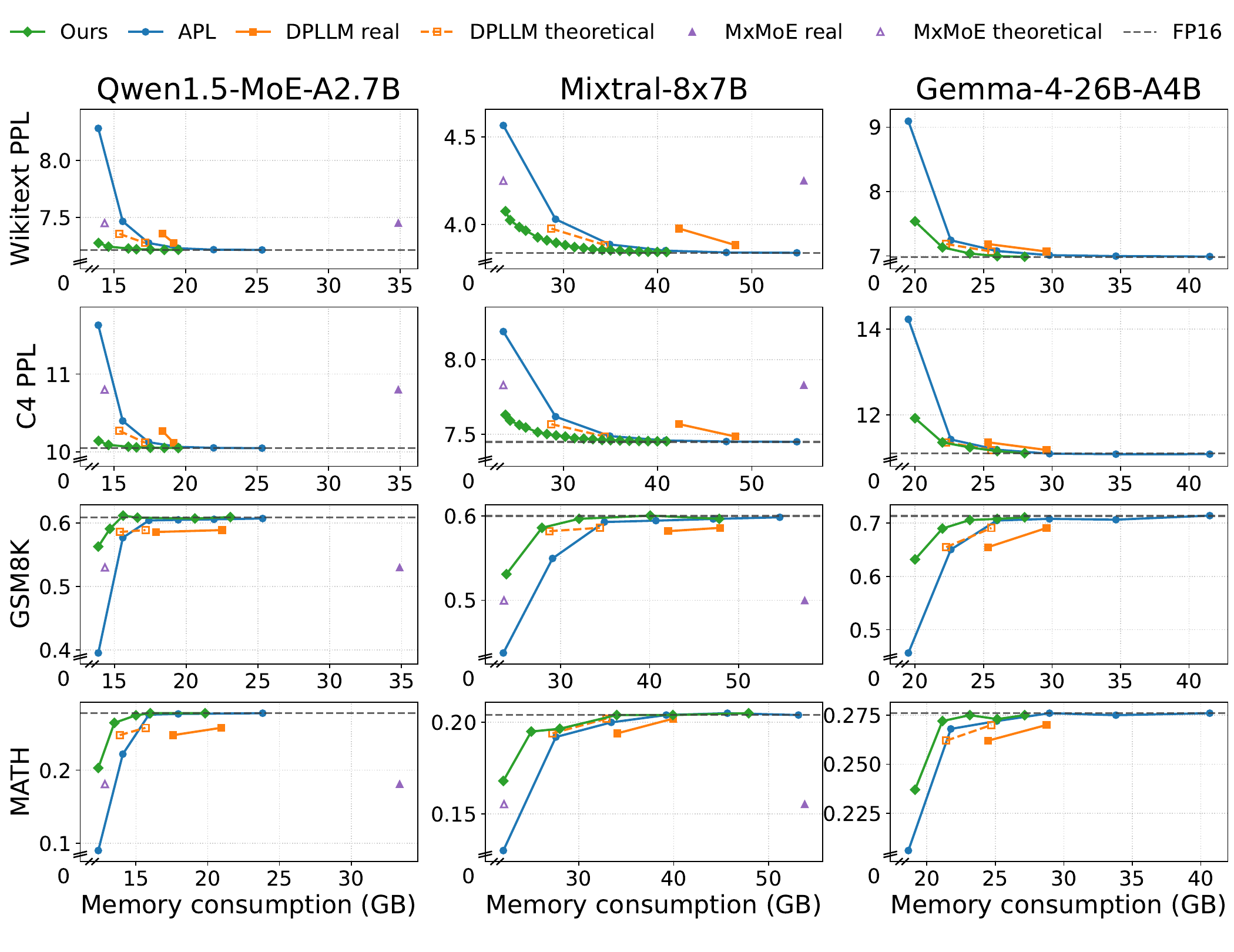}
\vspace{-.3in}
\caption{Quality-memory tradeoff across three MoE models and four evaluation tasks.}
\label{fig:quality-memory-tradeoff}
\end{figure}

Figure~\ref{fig:quality-memory-tradeoff} reports model
quality under different GPU memory consumption, the x-axis shows memory consumption in GB, and the y-axis shows task quality. 
We report perplexity on Wikitext2 and C4, where lower is better, and
reasoning accuracy on GSM8K and MATH-500, where higher is better.

\noindent\textbf{Baseline memory comparison.}
 For \NAME{} and APL, we report the measured real GPU memory consumption of the runtime.
For DP-LLM and MxMoE, \emph{theoretical} memory is the storage estimated directly from
the assigned bitwidths, while \emph{real} memory is the memory consumption
required by the existing implementation; we report both. DP-LLM needs to keep the higher-bit tensors in memory due to its dynamic
precision framework, and during inference, it dequantizes both the high and low-bit
precision paths. MxMoE uses ``fake'' quantization for its
weight-only quantization implementation, i.e.,
weights are simulated at low precision but stored and executed with
higher-precision tensors. It also 
 loads
multiple sets of weights, so its real memory consumption is much larger than the theoretical
memory consumption. We report MxMoE's 3.25-bit mixed-precision configuration, which gives the best
accuracy among its reported configurations.

\noindent\textbf{Overall Trends.}
Across all three models and tasks, \NAME{} always reaches high quality with the lowest memory consumption (Figure~\ref{fig:quality-memory-tradeoff}). These results show that selecting routed expert pages to reduce is more effective than using one static quantization precision policy.

\noindent$\bullet$ \textit{Perplexity Benchmarks: }
On Wikitext2 and C4, \NAME{} reaches the near-FP16 region at 16~GB on
Qwen1.5-MoE-A2.7B, 35~GB on Mixtral-8$\times$7B, and 22~GB on
Gemma-4-26B-A4B. APL loses more quality at the smallest memory consumption.
DP-LLM reaches similar quality only with higher theoretical and real memory consumption. MxMoE shows better quality than the uniform quantization with theoretical memory consumption, but performs still worse than \NAME{}.

\noindent$\bullet$ \textit{Reasoning Accuracy Benchmarks:}
GSM8K and MATH-500 show the same trend as perplexity. Under the same memory
consumption, \NAME{} consistently achieves higher accuracy than the baselines.
The improvement is more significant under tighter memory budgets.

\subsection{Long-Context Quality under Large KV-Cache Demand}
\label{sec:evaluation-long-context}

Long-context serving keeps a large KV cache on GPU. To answer RQ2, we evaluate whether \NAME{} preserves
long-context quality under large KV-cache demand. We use Qwen1.5-MoE-A2.7B and
three LongBench tasks: Passage Retrieval, NarrativeQA, and QMSum. \mbox{Higher scores
are better.}

\begin{table}[tbh]
  \centering
  \caption{Long-context quality under different memory consumption on Qwen1.5-MoE-A2.7B. Each row of APL uses approximately the same memory as the \mbox{corresponding row of \NAME{}.}}
  \label{tab:long-context-quality}
  \renewcommand{\arraystretch}{0.95}
  \begin{tabular}{lccccc}
    \toprule
    Method & Memory (GB) & Passage Retrieval & NarrativeQA & QMSum & Average \\
    \midrule
    FP16 & 35.25 & 15.5\% & 11.9\% & 23.5\% & 17.0\% \\
    \midrule
    APL-3bit & 9.40 & 10.5\% & 4.9\% & 21.1\% & 12.2\% \\
    APL-4bit & 11.11 & 14.0\% & 7.2\% & 22.9\% & 14.7\% \\
    APL-5bit & 12.92 & 13.0\% & 9.7\% & 23.4\% & 15.4\% \\
    APL-6bit & 14.97 & 15.5\% & 12.0\% & 23.5\% & 17.0\% \\
    \midrule
    \NAME-10GB & 9.86 & 17.5\% & 10.0\% & 23.6\% & 17.0\% \\
    \NAME-12GB & 11.82 & 14.5\% & 9.4\% & 23.3\% & 15.7\% \\
    \NAME-13GB & 12.79 & 15.0\% & 11.2\% & 23.4\% & 16.5\% \\
    \NAME-15GB & 14.83 & 15.5\% & 11.4\% & 23.6\% & 16.8\% \\
    \bottomrule
  \end{tabular}
  \vspace{-.1in}
\end{table}

\noindent\textbf{Overall Trends.}
Table~\ref{tab:long-context-quality} shows that \NAME{} preserves
long-context quality with much lower memory consumption than FP16. At the 10~GB budget, \NAME{} reaches the same 17.0\% average score as FP16 while using only 9.86~GB. In contrast,
APL reaches only 12.2\% at a similar memory consumption.
At medium memory consumption, \NAME{} continues to outperform APL. Under the 13~GB budget, \NAME{} reaches 16.5\% average score, while APL only reaches
15.4\%. The largest gains appear under tight memory, where \NAME{} improves Passage Retrieval from 10.5\% to 17.5\% and NarrativeQA from 4.9\% to 10.0\%.

\subsection{Serving Throughput}
\label{sec:evaluation-efficiency}

\begin{wraptable}{r}{0.5\linewidth}
  \centering
  \vspace{-.25in}
  \caption{Qwen1.5-MoE-A2.7B throughput.}
  \label{tab:decode-throughput}
  \footnotesize
  \setlength{\tabcolsep}{3pt}
  \renewcommand{\arraystretch}{0.9}
  \begin{tabular}{lrrrr}
    \toprule
    Method & B=1 Mem. & B=1 TPS & B=4 Mem. & B=4 TPS \\
    \midrule
    FP16 & 27.04 & 67.1 & 28.17 & 258.0 \\
    \midrule
    Uniform & 7.63 & 134.5 & 8.75 & 429.9 \\
    Uniform & 9.29 & 123.5 & 10.41 & 413.8 \\
    Uniform & 11.15 & 126.1 & 12.28 & 404.1 \\
    Uniform & 13.44 & 123.1 & 14.56 & 401.9 \\
    \midrule
    \NAME{} & 7.63 & 130.1 & 8.73 & 419.4 \\
    \NAME{} & 9.22 & 124.8 & 10.34 & 404.9 \\
    \NAME{} & 11.10 & 122.1 & 12.28 & 394.8 \\
    \NAME{} & 13.34 & 120.1 & 14.55 & 385.4 \\
    \bottomrule
  \end{tabular}
  \vspace{-.2in}
\end{wraptable}

This section reports generation throughput in tokens per second (TPS). To
answer RQ3, we evaluate whether \NAME{} maintains throughput while
using lower memory consumption. 

Table~\ref{tab:decode-throughput} reports long-decode throughput at sequence
length 2048 under batch sizes 1 and 4.
The memory columns report measured GPU
memory consumption, and the TPS columns report generation throughput.
Overall, \NAME{} matches the throughput of the uniform APL-style
baseline with similar or lower memory consumption. The largest throughput
drop is 3.3\% at batch size 1 and 4.1\% at batch~size~4.

\vspace{-.06in}
\subsection{Ablation Studies}
\label{sec:evaluation-ablations}
\vspace{-.06in}

\begin{wraptable}[8]{r}{0.42\linewidth}
  \centering
  \vspace{-.25in}
  \caption{Ablation study of \NAME{}.}
  \label{tab:qwen-ppl-ablation}
  \footnotesize
  \begin{tabular*}{\linewidth}{@{\extracolsep{\fill}}lcc@{}}
    \toprule
    Configuration & Wikitext2 PPL & C4 PPL \\
    \midrule
    \NAME{} & 7.22 & 10.06 \\
    w/o routing statistics & 7.26 & 10.13 \\
    w/o prompt residual & 7.31 & 10.19 \\
    w/o page movement & 7.43 & 10.33 \\
    w/o global sensitivity & 7.46  & 10.40  \\
    \bottomrule
  \end{tabular*}
\end{wraptable}
We isolate the effect of each component on Qwen1.5-MoE-A2.7B perplexity. Lower perplexity is better.
Table~\ref{tab:qwen-ppl-ablation} shows that the full \NAME{} system achieves
the lowest perplexity. Removing routing statistics or
prompt residuals degrades perplexity to 7.26/10.13 and 7.31/10.19,
showing that both online measurements help. Removing page movement turns
\NAME{} into a static mixed-precision plan and further increases perplexity to
7.43/10.33. Removing global sensitivity falls back to uniform \mbox{quantization and
performs worst.}

\vspace{-.04in}

\section{Related Work}
\label{sec:related-work}
\vspace{-.03in}

\noindent\textbf{LLM KV Cache Management.}
A key goal of LLM serving systems is to efficiently store and manage the KV cache
created during autoregressive generation.
Many systems feature some optimizations to better organize and de-fragment the KV cache,
such as vLLM's PagedAttention~\cite{vllm}, TTKV~\cite{ttkv}, LayerKV~\cite{xiong2024layerkv},
and DiffKV~\cite{diffkv}, or offload it to CPU or disk when GPU memory is insufficient
\cite{flexgen, kumar2025aqa, hybridgen}.
Some systems also compress the KV cache directly using quantization,
such as MoQAE and KVTuner~\cite{tao2025moqae,kvtuner},
or with lossy compression algorithms~\cite{diffkv,kvcomp}.
These methods optimize the memory consumption of the KV cache.
\NAME{} is complementary as it acts on MoE weights,
which creates additional available space for a growing KV cache.

\noindent\textbf{Post-Training and Runtime Model Quantization.}
Post-training quantization reduces model memory consumption and bandwidth demand.
Many existing works, such as GPTQ~\cite{gptq}, SmoothQuant~\cite{smoothquant},
AWQ~\cite{awq}, ARQ~\cite{yang2026arq}, and SpQR~\cite{spqr} explore various quantization strategies
to improve the amount of compression achieved while controlling accuracy loss.
These quantization methods produce a fixed quantized model before serving.
To meet runtime objectives, such as dynamic latency targets at serving time,
some works dynamically adjust bitwidths globally or layer-wise,
such as Any-Precision LLM~\cite{apl} and DP-LLM~\cite{dpllm}.
\NAME{} is a higher-level system that focuses on LLM memory management at serving time,
  adapting dynamic, mixed-precision quantization of MoE weights
as one of its key techniques to counter KV cache memory pressure.

\noindent\textbf{MoE Serving Systems and Expert-Aware MoE quantization.}
Mixture-of-Experts (MoE) architectures route input to a mix of multiple experts
\cite{gshard,switch,mixtral}.
Some recent systems optimize MoE serving cost by changing expert placement/residency
and prefetching~\cite{expertflow, klotski, moespeq, moelens, fluxmoe, megascaleinfer},
but are not capable of quantization.
Quantization is profitable on MoE, since the weights of experts can differ significantly
in quantization sensitivity,
which motivates \textit{expert-aware} quantization strategies.
Recent systems, including QuantMoE-Bench~\cite{quantmoebench},
MiLo~\cite{milo}, MxMoE~\cite{mxmoe}, EAQuant~\cite{fu2025eaquant}, and MoQE~\cite{moqe},
explore various such quantization strategies.

In parallel to our work, researchers proposed DynaExq and DyMoE~\cite{dynaexq, dymoe}, which are capable of
  dynamic MoE quantization.
  However, \NAME{} allows for much finer-grained control over submodules of experts
  compared to per-expert control in DynaExq and DyMoE.
  \NAME{} quantization is guided by real-time memory pressure from KV cache
  instead of a constant HBM usage cap assumed in DynaExq,
or the memory limit of the edge device in DyMoE.

\vspace{-.05in}

\section{Conclusion}
\vspace{-.05in}
\label{sec:con_lim}
We introduce \NAME{}, a runtime memory-management system for MoE LLM serving
under KV-cache pressure. \NAME{} treats Any-Precision expert bit-planes and LUTs
as weight pages, dynamically adjusts GPU-resident expert precision at safe
boundaries, and uses a quality-aware planner to select low-damage reductions.
Across three MoE models, \NAME{} improves the quality--memory tradeoff over baselines on language modeling, reasoning, and long-context tasks while
maintaining throughput close to the uniform baseline.

We demonstrated \NAME{}'s performance on specific quantization formats and sensitivity metrics. 
Future work includes extending \NAME{} to other quantized weight formats with
compatible layouts and kernels, and exploring complementary methods for
estimating prompt-wise expert sensitivity to further improve the
quality--memory tradeoff under KV-cache pressure.

\subsection*{Acknowledgments}

This research was supported in part by the NSF grants No. CCF-2217144 and CCF-2313028, and the IBM-Illinois
Discovery Accelerator Institute. This research used DeltaAI advanced computing and data resource,
supported by the NSF (award OAC 2320345) and the State of Illinois.

\bibliographystyle{plainnat}
\bibliography{ref}

\appendix

\end{document}